\documentclass{article}

\PassOptionsToPackage{numbers,compress}{natbib}
\usepackage{genai4health_2026}
\usepackage[normalem]{ulem}

\usepackage[utf8]{inputenc}
\usepackage[T1]{fontenc}
\usepackage{hyperref}
\usepackage{url}
\usepackage{booktabs}
\usepackage{amsfonts}
\usepackage{amsmath}
\usepackage{tabularx}
\usepackage{amssymb}
\usepackage{mathtools}
\usepackage{nicefrac}
\usepackage{censor}
\usepackage{microtype}
\usepackage{xcolor}
\usepackage{graphicx}
\usepackage{subcaption}
\usepackage[inline]{enumitem}
\usepackage{amsthm}
\usepackage{comment}

\theoremstyle{definition}

\theoremstyle{remark}

\usepackage{adjustbox}
\usepackage{makecell}
\usepackage{array}
\usepackage{titletoc}
\usepackage{float} 
\hypersetup{colorlinks=true,linkcolor=blue,citecolor=blue,urlcolor=blue}

\title{Sensorium: An Open-Source Multimodal Data Collection and Curation Library for Embodied Surgical AI}

\author{
  Lorenzo Mazza$^{1,\dagger}$\\
\texttt{lorenzo.mazza@nct-dresden.de}
\And
Ariel Rodriguez$^{1,2, \dagger}$\\
\texttt{ariel.rodriguezjimenez@nct-dresden.de}
\And
Stefanie Speidel$^{1,2 , 3,4}$
}

\begin{document}

\maketitle

\begingroup
\renewcommand{\thefootnote}{}
\footnotetext{%
$^\dagger$Corresponding authors. 
$^1$NCT/UCC Dresden, UKDD Dresden, TU Dresden, DKFZ Heidelberg.
$^2$BMFTR Research Hub 6G-Life. 
$^3$Cluster of Excellence CeTI. 
$^4$HZDR Dresden.%
}
\endgroup

\begin{abstract}
Embodied surgical AI requires reliable multimodal data, yet surgical data collection remains costly, difficult to repeat, and strongly dependent on the hardware setup. We present Sensorium, an open-source Python library for recording, processing, annotating, and curating multimodal surgical data from configurable robot and sensor streams using the Robot Operating System (ROS) and ROS~2 middleware. The library combines operator-facing data acquisition, configurable processing, semantic annotation, and episodic dataset generation. We evaluate Sensorium across 1{,}516 recordings of phantom, ex vivo, and in vivo porcine experiments involving multiple surgical tasks and sensing configurations, and report preliminary usability feedback.
\end{abstract}

\begin{figure}[H]
    \centering
    \includegraphics[width=0.7\linewidth]{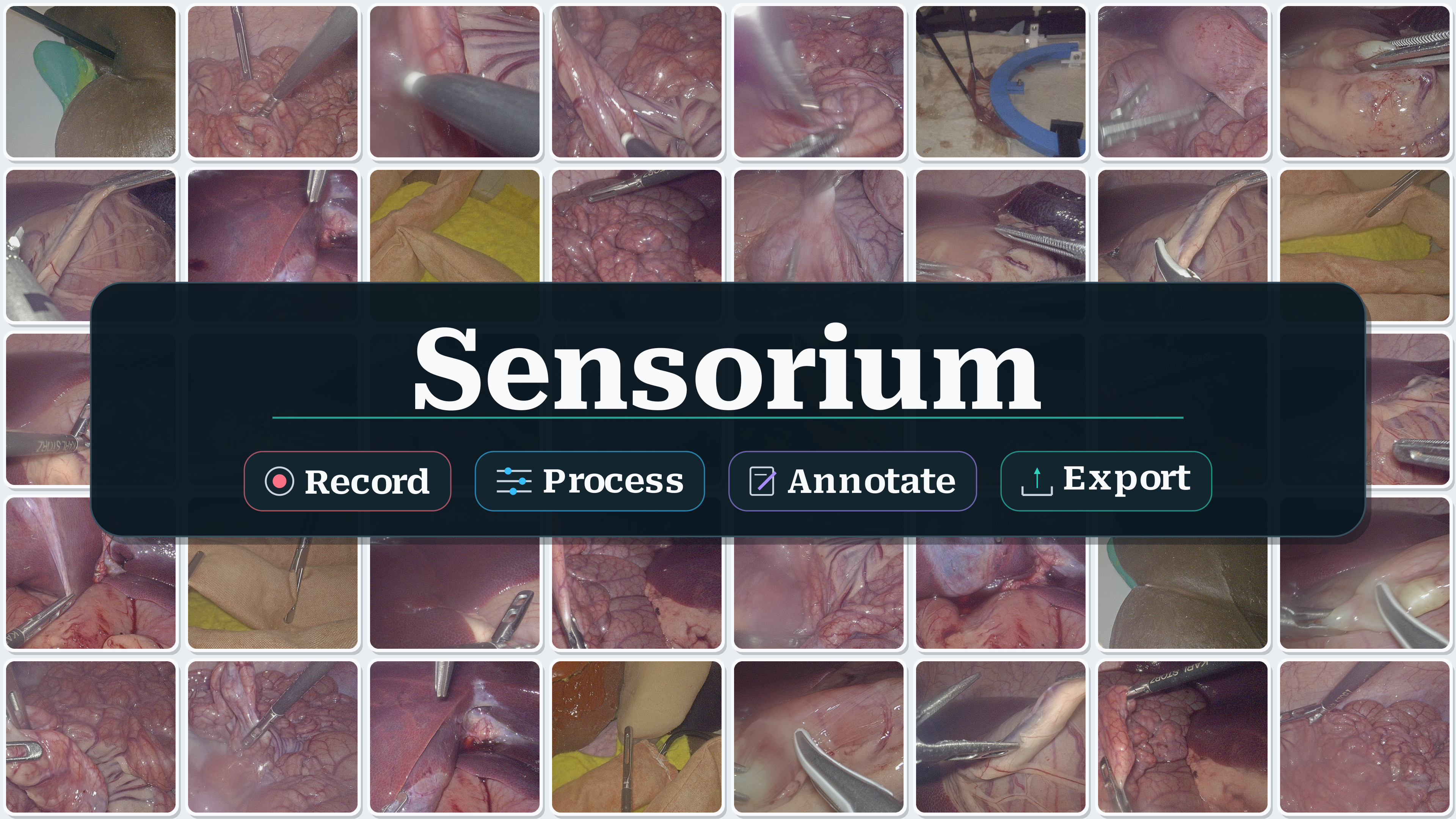}
    \caption{Sensorium provides a unified workflow for recording, processing, annotating, and exporting multimodal surgical data. The mosaic shows representative endoscopic views from the tasks and experimental settings summarized in Table~\ref{tab:deployment-coverage}.}
    \label{fig:recording-mosaic}
\end{figure}

\section{Introduction}

Recent progress in embodied surgical AI has increasingly relied on multimodal recordings of surgical activity, including video, robot kinematics, and semantic annotations. This trend is reflected in the emergence of multimodal surgical datasets~\cite{imitatecholec,openhembodiment2026}, which facilitate a range of learning paradigms and applications, from behavioral cloning for surgical task automation~\cite{kim2025srt,kim2025srth,haworth2025suturebot,mazza2026supervised,rodriguez2026larmoe,chen2026surgvil}, to surgical vision-language models~\cite{schmidgall2024gp,li2024llava,jin2024surgicalllava,zeng2025surgvlm,jiang2026surgr1} and generative surgical world models~\cite{he2026cosmoshsurgical,nvidiacosmoshdreams}.

However, acquiring such multimodal data and annotations in surgical settings is particularly expensive~\cite{dupont2025grandchallenges}. Data collection protocols require clinical input, clinicians have limited availability, and in vivo experiments
require substantial preparation, coordination, and ethical approval.
Recording tools must therefore support reliable acquisition and
preserve data for reuse with different representations, sampling
rates, and annotation protocols. Scarcity and limited repeatability also make usability a central requirement of the recording process: operators need clear indications of sensor readiness and recording state, along with immediate episode review before the experimental setup changes. The infrastructure should also accommodate varying robot and sensor interfaces across laboratories. This diversity is evident when looking at platforms used for surgical robotics research, including da Vinci Research Kit (dVRK)~\cite{kazanzides2014dvrk}, KUKA robotic platforms~\cite{shademan2016supervised}, Franka Emika Panda~\cite{kawaharazuka2024constrained}, and UR5/UR5e~\cite{schwaner2021mops, rodriguez2026testbed}, as well as in Open-H-Embodiment Dataset~\cite{openhembodiment2026}, which aggregates data from a large variety of commercial, research, and custom robotic platforms.

Recent open-source multimodal data acquisition systems address parts of these requirements. SurgSync~\cite{zhou2026surgsync} provides timestamp-aligned acquisition recording of robot kinematics, stereo endoscopic video, and additional sensing on the dVRK, together with tools for post-processing and annotation. Its acquisition stack and hardware integration are designed around the dVRK and its associated sensing setup. MiDAS~\cite{weerasinghe2026midas} is a framework targeted for robotic platforms that do not expose robot telemetry: it employs external electromagnetic and RGB-D tracking, and foot-pedal sensing to approximate robot kinematics. This leaves a gap for a general-purpose framework that directly records heterogeneous robot and sensor data while supporting their processing and annotation for embodied surgical AI systems.

To address this gap, we present Sensorium, an open-source Python library for collecting, processing, annotating, and curating multimodal surgical data for embodied AI applications. Sensorium integrates custom sensor modalities through ROS and ROS 2, and provides both a command-line interface and a browser-based graphical interface (GUI) for local and remote operation. The library includes tools for timestamp-based temporal alignment and stereo rectification, together with annotation functionalities for surgical phases~\cite{twinanda2016endonet,czempiel2020tecno} and language instructions for language-conditioned surgical robot learning~\cite{kim2025srth,haworth2025suturebot}. Additionally, Sensorium organizes recordings into reusable episodic datasets and supports both raw CSV-based representations and the widely adopted LeRobot format~\cite{cadene2026lerobot, openhembodiment2026}, facilitating their use across different surgical AI pipelines, including policy learning, vision-based force estimation, and world model training.  Sensorium has been deployed as the recording application of an open-source laparoscopic robotics testbed~\cite{rodriguez2026testbed}, where it was used to collect a dataset of surgical robot demonstrations for bowel grasping and retraction on phantom and ex vivo~\cite{mazza2026supervised, wang2026exploring}. It was also used to record two in vivo porcine datasets covering endoscope guidance, tissue grasping, and tissue retraction, which were subsequently included in~\cite{openhembodiment2026}.

 In summary, our contributions are the following:
\begin{itemize}[nosep]
\item An open-source, flexible, configuration-driven library for collecting and curating multimodal recordings from heterogeneous ROS-integrated surgical sensors and robots. 
\item An operator-oriented browser GUI that unifies recording supervision, episode review, surgical-phase and language annotation, and dataset generation. 
\item A real-world validation of the library across phantom, ex vivo, and in vivo porcine data-collection settings, complemented by preliminary usability feedback.
\end{itemize}
The library code is available at \footnote{https://github.com/nct-tso-robotics/tso\_sensorium}.
 \section{System Overview}

\begin{figure}[H]
    \centering
    \includegraphics[width=\linewidth]{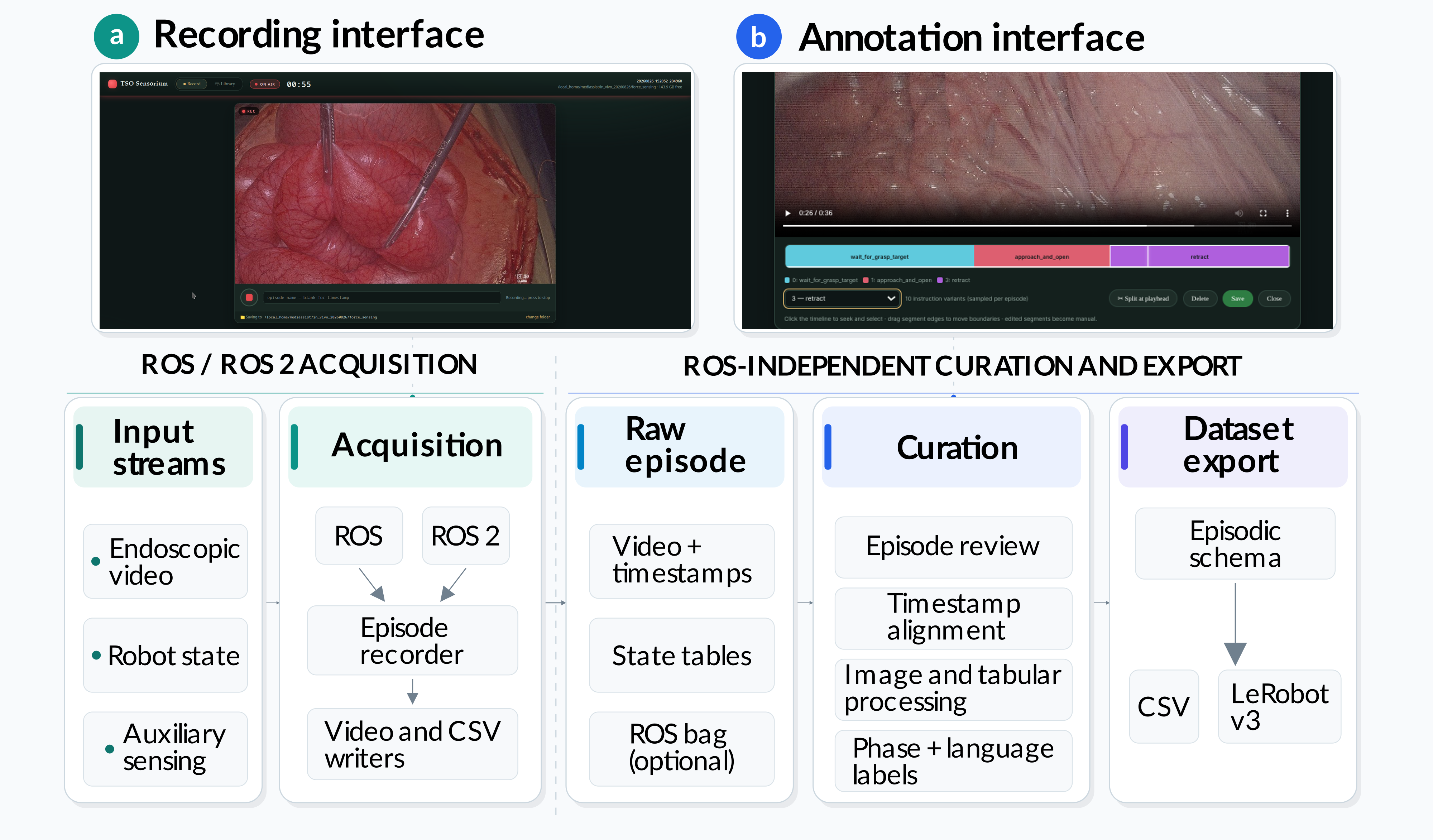}
      \caption{
          Overview of the Sensorium workflow.
          The browser interfaces support (a) multimodal recording and
          (b) episode review and semantic annotations.
          Recorded streams are stored as raw episodes, processed
          through a ROS-independent curation pipeline, and exported as episodic
          CSV or LeRobot v3 datasets.
      }
      \label{fig:system-workflow}
\end{figure}

Sensorium organizes multimodal surgical data collection around episodes, as illustrated in Figure~\ref{fig:system-workflow}. Acquisition runs alongside the robotic system through ROS and stores each recording as a raw episode containing timestamped video and state streams. Subsequent curation, processing, annotation, and dataset export are ROS-independent and can therefore be performed later on a separate workstation. Validated configuration files define both the streams acquired during recording and how raw episodes are subsequently processed and exported.

\subsection{Multimodal Acquisition}
Sensorium acquires heterogeneous data streams through the ROS middleware. Any device that exposes its measurements as a ROS topic can be added through configuration, without modifying the recording core. Across our deployments, this enabled acquisition from stereo endoscopes, pedal boards, haptic input devices, force sensors, and robotic platforms, as well as calibrated robot--camera transformations.
In our deployments, the clocks of all workstations running the robot,
camera, and other sensors are synchronized using the Precision Time
Protocol (PTP). The recorder always preserves the publication timestamp present in each ROS message header. Complete ROS bag recording can optionally be enabled to retain the original message streams. Starting a recording creates a self-contained raw episode containing the selected video and state streams.  The browser recording interface, shown in Figure~\ref{fig:system-workflow}a, provides a live camera view and displays the availability of configured sensor streams, the recording state, and the elapsed episode time. Before acquisition begins, required streams and the output location are checked, allowing recording failures to be detected while the experimental setup can still be inspected or corrected. Once an episode is stopped, it is finalized and becomes immediately available for review.
\subsection{Curation and Processing}
The curation interface in Figure~\ref{fig:system-workflow}b can run either as part of the recording application or independently over an existing recording directory. Users can browse video streams and divide an episode into temporal segments using a dataset-configurable phase-to-language mapping. Each phase can be associated with multiple paraphrased language instructions, increasing textual diversity in the recorded data. A single segment or dataset-level task description can instead provide a global instruction for coarse-grained tasks. These annotations provide aligned textual and categorical phase annotations, supporting language-conditioned policy learning~\cite{haworth2025suturebot}, surgical VLM training~\cite{jin2024surgicalllava, chen2026surgvil,li2024llava}, and surgical phase supervision~\cite{czempiel2020tecno, twinanda2016endonet, mazza2026supervised}.
For data processing, selected streams are first restricted to their common temporal interval and aligned to a configurable reference stream using nearest-neighbor timestamp matching. For each reference timestamp \(t_i\), the closest sample from stream \(k\) is selected as \[ j_i^{(k)}= \operatorname*{arg\,min}_{j} \left|t_i-t_j^{(k)}\right|. \] Episodes exceeding a configurable alignment tolerance are rejected. Configurable transformations can then be applied to the aligned data, including stereo rectification, resizing, and coordinate-frame transformations. Because these operations are performed after acquisition, different representations can be generated from the same raw episode. 
\subsection{Dataset Construction and Export}
\begin{figure}[]
    \centering
    \includegraphics[width=0.8\linewidth]{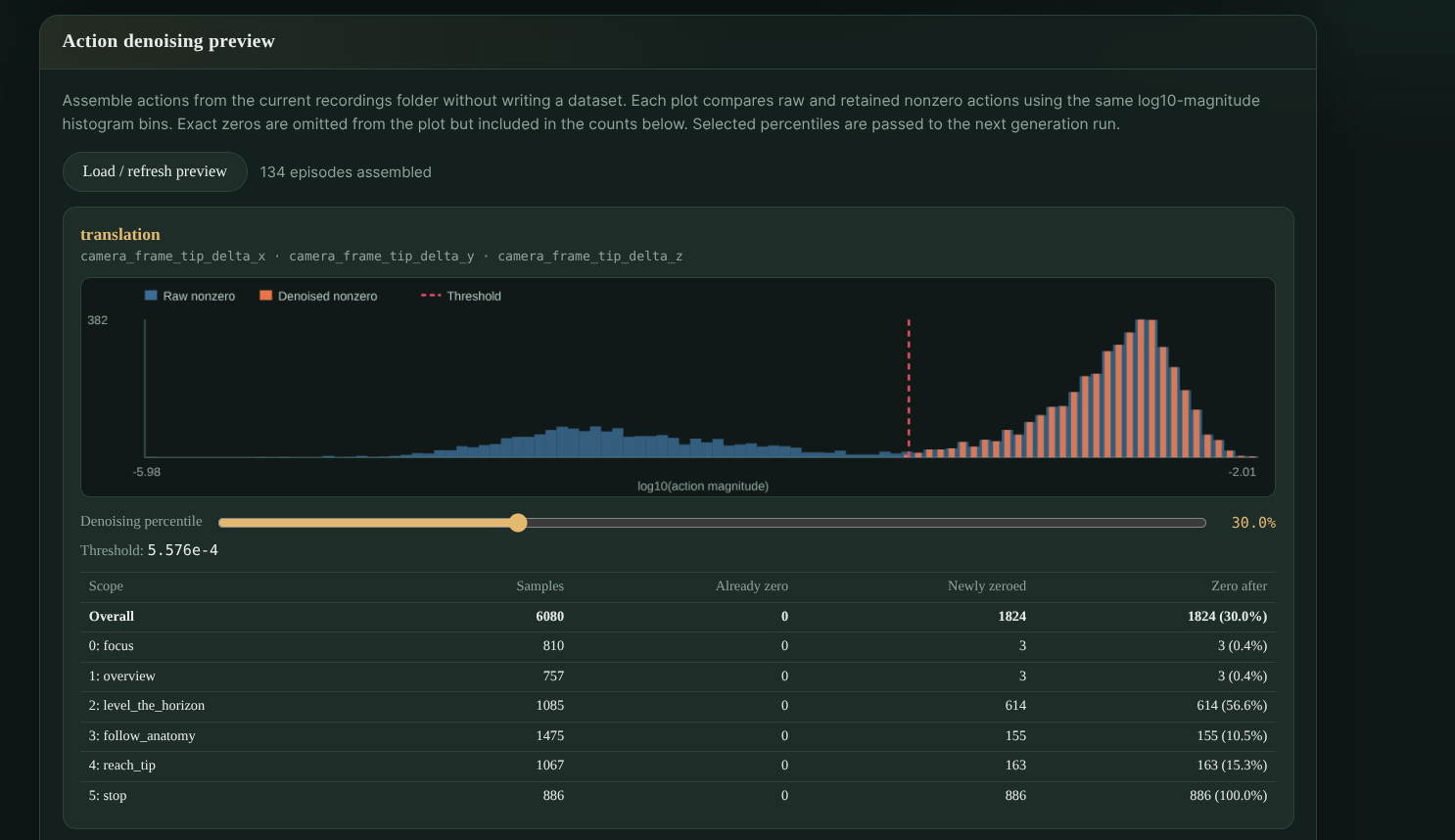}
    \caption{Action-denoising preview in the browser interface. Blue and
    orange bars compare raw and retained nonzero translation-action
    magnitudes on a \(\log_{10}\) scale. The dashed line marks the selected
    30th-percentile threshold. The table summarizes the samples that would
    be set to zero, overall and by task phase.}
    \label{fig:action-denoising}
\end{figure}

After alignment and processing, a configurable episodic schema defines the observations, robot states and actions, and textual/categorical annotations included in the final dataset. Phase and language annotations are projected onto the aligned timeline and stored alongside the selected sensor and robot data. Sensorium currently supports two export formats: per-episode CSV tables with optional losslessly encoded PNG frames or LeRobot v3 datasets containing Parquet-backed features and image or video observations~\cite{cadene2026lerobot}. For LeRobot export, actions can be selected from recorded sensor streams or derived from observations, with optional magnitude-based filtering to suppress small action fluctuations (Figure~\ref{fig:action-denoising}). Appendix~\ref {app:configuration-example} provides an example configuration, including observation/action mappings and filtering thresholds. Each episode is validated independently before export, so recordings that do not satisfy the configured alignment or schema requirements are excluded and reported without interrupting construction of the remaining dataset.

\paragraph{Demonstration plan.}
We will demonstrate episode review, surgical-phase and language annotation, action filtering, and LeRobot export through Sensorium's browser interface using previously collected surgical recordings. A prerecorded video will illustrate data acquisition on the robotic platform, allowing the complete workflow to be presented without
requiring surgical hardware at the workshop.

\begin{table}
    \centering
    \caption{Deployment coverage and post-recording temporal alignment.
    Durations are rounded to the nearest minute. Alignment statistics
    summarize pooled nearest-neighbor residuals between reference-camera
    and robot-state timestamps over their common temporal interval.}
    \label{tab:deployment-coverage}
    \label{tab:temporal-alignment}

    \scriptsize
    \setlength{\tabcolsep}{3.5pt}
    \renewcommand{\arraystretch}{1.10}

    \begin{tabularx}{\linewidth}{
        @{}>{\raggedright\arraybackslash}X r l r r r@{}
    }
        \toprule
        Dataset & Episodes & Streams & Duration
            & \multicolumn{2}{c}{Timestamp residual (ms)} \\
        \cmidrule(lr){5-6}
        & & & (h:min) & Median & 95th \\
        \midrule
        Bowel grasping/retraction demonstrations (phantom)
            & 180 & V,R,G,T,P,F,I & 1:41 & 8.60 & 17.68 \\
        Bowel grasping/retraction demonstrations (ex vivo)
            & 123 & V,R,G,T,F,I & 0:59 & 8.13 & 16.65 \\
        Bowel grasping/retraction demonstrations (in vivo)
            & 66 & V,R,G,T,P,F,I & 0:31 & 8.10 & 16.65 \\
        Bowel grasping/retraction policy rollouts (mixed)
            & 426 & V,R,G,T,P & 4:44 & 8.10 & 16.62 \\
        Endoscope guidance (phantom)
            & 276 & V,R,T,P,G & 2:10 & 6.41 & 16.32 \\
        Endoscope guidance (in vivo)
            & 122 & V,R,P,T & 1:13 & 8.08 & 16.56 \\
        Liver lifting (phantom)
            & 30 & V,R,T & 0:12 & 0.82 & 15.58 \\
        Multimodal bowel grasping demonstrations (phantom)
            & 90 & V,R,G,T,P & 0:08 & 8.12 & 16.58 \\
        Multimodal bowel grasping policy rollouts (phantom)
            & 107 & V,R,G,T,P & 0:29 & 7.93 & 16.51 \\
        Thread pulling/retraction (in vivo)
            & 49 & V,R,G,T,P & 0:32 & 7.91 & 16.53 \\
        Tissue displacement (in vivo)
            & 43 & V,R,G,T,P & 0:13 & 0.49 & 0.99 \\
        Open-surgery bowel retraction (in vivo)
            & 4 & V,R,T,F,I & 0:12 & 7.78 & 16.40 \\
        \midrule
        Total & 1{,}516 & -- & 13:04 & -- & -- \\
        \bottomrule
    \end{tabularx}

    \smallskip
    {\raggedright\footnotesize
    V: stereo video; R: robot state; G: gripper;
    T: camera--robot transform; P: pedal; F: force; I: IMU.
    The combined stereo stream was used as reference when available;
    256 rollout recordings instead used the left stream.
    \par}
\end{table}
\section{System Evaluation}
We evaluate Sensorium along three dimensions: deployment breadth across surgical tasks and experimental settings, temporal consistency of the constructed episodes, and preliminary usability for experiment operators.
\paragraph{Deployment breadth and downstream use.} Overall, the system was used for 1{,}516 recordings using the UR5e robotic platform from \cite{rodriguez2026testbed}, spanning demonstrations and policy rollouts in phantom, ex vivo, and in vivo \footnote{\tiny All procedures were approved by the local state authority (redacted for anonymity) 
and conducted in accordance with institutional ethical standards for animal experimentation and the registered protocol.} settings. Table~\ref{tab:deployment-coverage} summarizes the recorded cohorts
and sensing modalities: the recordings comprise 13~h~04~min of data across bowel grasping and retraction, endoscope guidance, liver lifting, thread pulling, tissue
displacement, and open-surgery bowel retraction. They include combinations of
stereo video, robot state, gripper state, pedal input, force sensing, IMU
measurements, and camera--robot transformations. As a downstream validation, datasets collected and curated using Sensorium were used to train the behavioral cloning policies evaluated in \cite{mazza2026supervised, rodriguez2026larmoe} and the
vision-based force-estimation models evaluated in~\cite{wang2026exploring}. Recordings have also been used for preliminary world-model training in ongoing research. This demonstrates the complete pipeline from heterogeneous hardware to surgical AI training data.
\paragraph{Temporal alignment.} We evaluated the timestamp-alignment procedure by measuring the nearest-neighbor
residual between each reference-camera timestamp and its associated sensor-state
samples. As shown in Table~\ref{tab:temporal-alignment}, the median residual was
at most 8.60~ms across all cohorts, while the 95th percentile remained at most
17.68~ms.
\paragraph{Preliminary operator usability.} Additionally, seven users of Sensorium provided preliminary feedback through an eight-item
5-point Likert questionnaire~\cite{likert1932technique} (1: strongly disagree, 5: strongly agree). Overall
ease of use, integration into the procedural workflow, visibility of sensor
readiness, and willingness to use the system in future recording sessions all
received a mean score of 5.0/5. Starting and stopping recordings, identifying
when recording was active, and confidence that data were saved received
4.86/5.  Open-ended responses highlighted the simplicity of the recording interface,
rapid access to recorded episodes, and remote operation. Suggested improvements
included recording shortcuts, clearer feedback during saving, and additional
separation between episode-deletion controls. More details are presented in Appendix \ref{app:user_feedback}.

\section{Conclusion}

We presented Sensorium, an open-source Python library for recording data from ROS-based sensors and devices, annotating and curating recordings, and exporting them into datasets ready for downstream AI applications, including policy learning, world models, and vision-based force estimation.
Limitations include the need to wrap sensors without native ROS support in ROS nodes, dataset export being tied to the selected reference stream without support for resampling to a user-defined frequency, and a user evaluation limited to seven participants from a single laboratory, restricting the generalizability of the usability findings to other workflows and hardware configurations.
Future work will focus on evaluating Sensorium across different robotic platforms, experimental setups, and research sites, and on extending the library based on feedback from the broader community. We hope that its open-source release facilitates wider adoption and contributes to more reproducible and accessible data collection workflows for robotics and AI.

\begin{ack}
Lorenzo Mazza and Stefanie Speidel acknowledge support by the project ”Next Generation AI Computing (gAIn),” funded by the Bavarian Ministry of Science and the Arts and the Saxon Ministry for Science, Culture, and Tourism as well as by the Hightech Agenda Bavaria. 

Ariel Rodriguez and Stefanie Speidel acknowledge the support by the German Research Foundation (DFG, Deutsche Forschungsgemeinschaft) as part of Germany’s Excellence Strategy – EXC 2050/1 – Project ID 390696704 – Cluster of Excellence “Centre for Tactile Internet with Human-in-the-Loop” (CeTI) of Technische Universität Dresden.

Ariel Rodriguez and Stefanie Speidel would additionally like to thank the Federal Ministry of Research, Technology, and Space (BMFTR) for its support as part of the research program Communication Systems “Souverän. Digital. Vernetzt”. Joint project 6G-life, project identification number: 16KIS2413K. 

This work has been partially supported by the German Federal Ministry of Research, Technology and Space (BMFTR) under the Robotics Institute Germany (RIG).
\end{ack}
\clearpage
\bibliographystyle{plainnat}
\bibliography{references}


\appendix

\section{User Feedback}
\label{app:user_feedback}
Seven users from a single laboratory completed an eight-item questionnaire
using a 5-point Likert scale (1: strongly disagree; 3: neither agree nor
disagree; 5: strongly agree). Table~\ref{tab:user-feedback} reports the
responses using anonymous respondent identifiers. All seven respondents
consented to quotation of their anonymized feedback in a publication.

\begin{table}[H]
    \centering
    \caption{Individual operator usability ratings and item means. Questionnaire
    statements are reproduced as presented to respondents.}
    \label{tab:user-feedback}
    \small
    \setlength{\tabcolsep}{3pt}
    \renewcommand{\arraystretch}{1.15}
    \begin{tabular*}{\linewidth}{@{}>{\raggedright\arraybackslash}p{0.52\linewidth}@{\extracolsep{\fill}}*{7}{c}r@{}}
        \toprule
        Questionnaire item & R1 & R2 & R3 & R4 & R5 & R6 & R7 & Mean \\
        \midrule
        I could clearly determine whether all required sensors were ready
            & 5 & 5 & 5 & 5 & 5 & 5 & 5 & 5.00 \\[3pt]
        Starting and stopping a recording was straightforward
            & 5 & 5 & 5 & 5 & 4 & 5 & 5 & 4.86 \\[3pt]
        I could clearly tell when recording was active
            & 5 & 4 & 5 & 5 & 5 & 5 & 5 & 4.86 \\[3pt]
        The system fit into the procedural workflow without unnecessary disruption
            & 5 & 5 & 5 & 5 & 5 & 5 & 5 & 5.00 \\[3pt]
        I was confident that the intended data had been saved correctly
            & 5 & 5 & 5 & 5 & 4 & 5 & 5 & 4.86 \\[3pt]
        I could understand and respond to warnings or sensor failures
            & 3 & -- & 5 & 5 & 3 & 5 & 5 & 4.33 \\[3pt]
        I would use Sensorium for a future recording session
            & 5 & 5 & 5 & 5 & 5 & 5 & 5 & 5.00 \\[3pt]
        Overall, Sensorium was easy to use
            & 5 & 5 & 5 & 5 & 5 & 5 & 5 & 5.00 \\
        \bottomrule
    \end{tabular*}

    \smallskip
    \raggedright\footnotesize
    -- denotes an unanswered item. Means use the available responses:
    $n=7$ for all items except warnings or sensor failures ($n=6$).
    No responses were imputed or excluded on the basis of their score.
\end{table}

\paragraph{Error-message ratings.}
\uline{Both respondents who selected 3 had experienced no recording errors.}
Their neutral ratings therefore reflect the absence of an opportunity to
assess error messages during recording.

\paragraph{Open-ended feedback.}
Respondents highlighted the straightforward interface, accessible recording
list, ease of starting and stopping recordings, rapid removal of unwanted
episodes, and access to recording and review from devices on the local
network. One response identified slight uncertainty during saving. Suggestions
included a keyboard shortcut, voice control for starting and stopping
recordings, and greater spacing between episode rows to reduce accidental
deletion. One respondent reported needing a brief explanation of the
interface; four reported no required assistance, and two left that question unanswered. These observations are preliminary feedback from a
small, single-laboratory sample.

\clearpage
\section{In Vivo Deployment}
\label{app:invivo-deployment}
\begin{figure}[H]
    \centering
    \includegraphics[width=\linewidth]{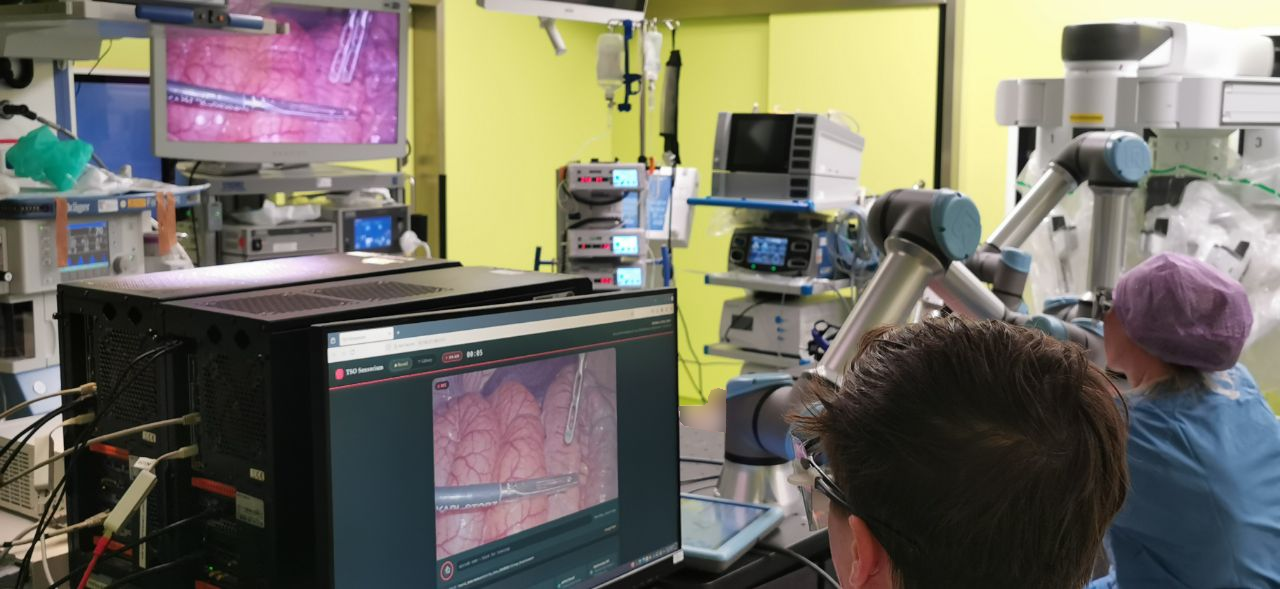}
    \caption{Sensorium during an in vivo porcine experiment. The foreground
    monitor shows the browser recording interface with an endoscopic view,
    while the surgical monitor displays the operative scene.}
    \label{fig:invivo-deployment}
\end{figure}

Figure~\ref{fig:invivo-deployment} shows Sensorium in use during an
in vivo porcine recording session, alongside the robotic platform and
surgical imaging equipment.

\clearpage
\section{Configuration Pseudocode}
\label{app:configuration-example}
The following pseudocode summarizes the endoscope guidance
configuration, showing its selected topics and observation/action mappings.
Executable configurations are included in the accompanying codebase ZIP.
The export names below follow the LeRobot writer.

{\footnotesize
\begin{verbatim}
record:
  stereo   <- "/stereo/camera_driver/image_raw"
  robot    <- "/ur5e_rcm_twist_controller/RobotState"
    fields: relative_tip_position, relative_pivot_RPY
  rotation <- "/robot_camera_transform"
    fields: transform.rotation.{x, y, z, w}

for each episode:
  align streams by nearest timestamp over their common interval
    reference: stereo; tolerance: 0.1 seconds
  left, right <- deinterlace, resize(960, 540), rectify(stereo)
  phase, language <- join annotations with the configured phase legend
  displacement <- xyz(robot.relative_tip_position)
  roll <- first_component(robot.relative_pivot_RPY)
  R <- rotation_matrix(rotation)  # camera-to-robot-base rotation

  for i = 0, ..., number_of_aligned_rows - 2:
    state[i] <- concat(inverse(R[i]) * displacement[i], roll[i])
    translation[i] <- inverse(R[i]) *
                      (displacement[i+1] - displacement[i])
    roll_change[i] <- wrap_to_[-pi, pi)(roll[i+1] - roll[i])
  discard the terminal row, which has no successor

over all successfully assembled episodes:
  zero translation vectors below the 30th percentile of nonzero norms
  zero roll changes below the 20th percentile of nonzero magnitudes

export each retained row i:
  observation.images.left  <- left[i]              RGB[540, 960, 3]
  observation.images.right <- right[i]             RGB[540, 960, 3]
  observation.state <- state[i]                    float32[4]
  action <- concat(translation[i], roll_change[i])  float32[4]
  task_phase <- phase[i]                           int64[1]
  task metadata <- language[i] or the default task description
\end{verbatim}
}

The state vector contains the recorded relative tip displacement expressed
in the current camera frame, followed by the recorded roll. Actions are
derived next-sample translation and wrapped roll increments. Denoising changes only the action components without overwriting raw data, and its computed thresholds are saved in dataset metadata.


\newpage

\end{document}